\documentclass[sigconf]{acmart}
\usepackage{multirow}
\AtBeginDocument{%
  \providecommand\BibTeX{{%
    \normalfont B\kern-0.5em{\scshape i\kern-0.25em b}\kern-0.8em\TeX}}}

\def\AAdel#1{\bgroup\markoverwith{\textcolor{blue}{\rule[0.5ex]{2pt}{1pt}}}\ULon{#1}}

\def\MSdel#1{\bgroup\markoverwith{\textcolor{red}{\rule[0.5ex]{2pt}{1pt}}}\ULon{#1}}

\begin{document}

\title{Zero-Shot Multi-Lingual Speaker Verification in Clinical Trials}

% \author{Ali Akram, Malikeh Ehghaghi, Marija Stanojevic, Jekaterina Novikova}
% \email{{akramsystems, malikehehghaghi, mstanojevic118, novikova.jekaterina}@gmail.com}
% \orcid{-, 0009-0000-9138-6444, 0000-0001-8227-6577, 0000-0003-4754-6126} # unfortunately, this doesn't work in this format; I'd need to split authors

\author{Ali Akram}
\email{akramsystem@gmail.com}
\affiliation{%
  \institution{Winterlight Labs, Cambridge Cognition}
  \country{Toronto, Canada}
}

\author{Marija Stanojevic}
\email{mstanojevic118@gmail.com}
\orcid{0000-0001-8227-6577}
\affiliation{%
  \institution{Winterlight Labs, Cambridge Cognition}
  \country{Toronto, Canada}
}

\author{Malikeh Ehghaghi}
\email{malikehehghaghi@mail.utoronto.ca}
%\orcid{0009-0000-9138-6444}
\affiliation{%
  \institution{Winterlight Labs, Cambridge Cognition}
  \country{Toronto, Canada}
}

\author{Jekaterina Novikova}
\email{novikova.jekaterina@gmail.com}
\orcid{0000-0003-4754-6126}
\affiliation{%
  \institution{Winterlight Labs, Cambridge Cognition}
  \country{Toronto, Canada}
}

\renewcommand{\shortauthors}{Akram et al.}

\begin{abstract}
Due to the substantial number of clinicians, patients, and data collection environments involved in clinical trials, gathering data of superior quality poses a significant challenge. In clinical trials, patients are assessed based on their speech data to detect and monitor cognitive and mental health disorders. We propose using these speech recordings to verify the identities of enrolled patients and identify and exclude the individuals who try to enroll multiple times in the same trial. Since clinical studies are often conducted across different countries, creating a system that can perform speaker verification in diverse languages without additional development effort is imperative. We evaluate pre-trained TitaNet, ECAPA-TDNN, and SpeakerNet models by enrolling and testing with speech-impaired patients speaking English, German, Danish, Spanish, and Arabic languages. Our results demonstrate that tested models can effectively generalize to clinical speakers, with less than 2.7\% EER for European Languages and 8.26\% EER for Arabic.  This represents a significant step in developing more versatile and efficient speaker verification systems for cognitive and mental health clinical trials that can be used across a wide range of languages and dialects, substantially reducing the effort required to develop speaker verification systems for multiple languages. We also evaluate how speech tasks and number of speakers involved in the trial influence the performance and show that the type of speech tasks impacts the model performance.
% TODO: ADD FIGURES TO PAPER
\end{abstract}




% \received{2 June 2023}
% \received[revised]{2 June 2023}
% \received[accepted]{4 August 2023}

\maketitle

\section{Introduction}

Extensive clinical trials span numerous patients, doctors, clinics, and even countries, making it hard to know if an enrolled patient is unique. They are also usually done over a long period of time, requiring participants to visit the doctor's office multiple times. \cite{shiovitz2013cns} found that 7.78\% of patients participating in large clinical trials were duplicated across different sites. This diminishes quality of results of billion dollars worth drug development process. We propose using machine learning models for speaker verification (SV) to solve this problem in cognitive and mental health trials that record speech data for assessment.

Speaker verification is the process of comparing unknown speech utterances to speech signals belonging to known enrolled individuals to decide if a new recording belongs to the known users. SV systems are used across industries including such as banking \cite{ramya2022enhanced, chen2009speaker}, transportation \cite{tamoto2019voice}, telecommunications \cite{melin1996speaker}, and healthcare \cite{arasteh2022speech, upadhyay2022feature}.

There are two types of SV systems: 1) text-dependent (TD), requiring the speaker to use the same text in enrollment and verification, and 2) text-independent (TI), without those constraints. TD systems dominated in the past because it was easier to verify users on repeated text \cite{irum2019speaker}. Recently, TI SV models are more common in practice since they only require audio, and there is no need for transcripts. TI models are comparable in accuracy with TD models \cite{chojnacka2021speakerstew} when pre-trained on large amounts of audio samples \cite{tu2022survey}. However, when it comes to cross-lingual SV, obtaining enough training data for each language is difficult, especially for clinical studies in low-resource languages.

Traditionally, TI SV systems depended fairly strongly on the language being spoken \cite{kleynhans2005language} and required language-specific training or fine-tuning. However, training and deploying a separate model for each language is inefficient, increasing the operational costs and overheads in multilingual systems. These concerns are even more substantial in clinical settings when working with patients with abnormal speech patterns.

We propose using large pre-trained SV TI model pre-trained on speech from multiple languages to verify speakers in clinical trials in zero-shot settings. Our target data comes from speakers with abnormal speech patterns. We also test the performance of those models on the speech recordings across different speech tasks and characterize the relation between the quality of SV and data properties (i.e. speech tasks and number of speakers).

\begin{flushleft}
In summary, the main contributions of this work are:
\end{flushleft}
\begin{itemize}
\item We propose using pre-trained end-to-end SV models to enroll and verify patients in cognitive and mental health clinical trials in zero-shot settings.
\item We test those models on the speech of English, German, Danish, Spanish, and Arabic patients and demonstrate the models can effectively generalize to different languages and speech impairment, achieving high performance as well as helping solve the duplicate participant problem.
\item We show how the performance changes when speech from different speech tasks is used, showing that speech task may influence EER.
\end{itemize}

\section{Related Work}
% Multiple references should be given in the first \cite
For a long time, SV models used a Gaussian mixture model universal background model (GMM-UBM) approach, where acoustic features (e.g., Mel frequency cepstral coefficients (MFCC)) were used to generate GMM models for each speaker from a set of speech recordings \cite{irum2019speaker}. The GMM-UBM-based models are easily trained, require a small number of speakers, and are often more interpretable than other approaches. However, GMM-UBM-based models assume the normal distribution of the data, which limits their ability to learn more complex speaker representations. The accuracy of SV models is evaluated using Equal Error Rate (EER) \cite{cheng2004method}, which is the point at which the true positive rate (TPR) is equal to the true negative rate (TNR). The GMM-UBM approach, as detailed in \cite{ozbek2018recognition}, achieved an Equal Error Rate (EER) of approximately 9.25\% on English datasets. However, those models are phoneme-dependent, making knowledge transfer between languages very challenging \cite{irum2019speaker}.

In recent years, significant advancements have been made in pre-trained deep-learning speaker verification (SV) models, achieving remarkably low EER of 0.68\% for English language \cite{koluguri2022titanet}. These models leverage the benefits of end-to-end semi-supervised learning, utilizing more readily available data in languages like English to generate improved speaker embeddings. Moreover, such models have demonstrated the ability to learn intricate and sophisticated representations. For instance, a neural-network-based architecture successfully learned the SV task along with language detection \cite{zhou2021language}. However, they tend to be less interpretable and require substantial computational resources and extensive datasets for training \cite{li2022interpretable}.

More recently, work has been done in assessing whether neural network-based models trained on a single source language (i.e., English) can be adapted to perform SV in another (target) language without the need to be fine-tuned or trained in that target language in zero-shot settings. For example, convolutional time-delay deep neural network (CT-DNN) \cite{li2017cross} architecture trained on the Fisher dataset \cite{cieri2004fisher}, containing 5000 English speakers, results in a model that can achieve EER of 3.71\% at the task of TI SV in Chinese/Uyghur. While we also use zero-shot adaptation, we use larger pre-trained models, and our target data comes from speakers with abnormal speech patterns.

%Google's SpeakerStew paper 
\cite{chojnacka2021speakerstew} shows that SV models trained on one language can be used for SV in other languages without requiring language-specific fine-tuning by testing 46 different languages in zero-shot settings. However, the datasets, which were used for training and evaluation in \cite{chojnacka2021speakerstew}, contained specific keywords (i.e., `Hey Google' and `Ok Google'), making the task partially TD. Their results might not generalize to speech that is unconstrained and fully TI. In addition, their model is also only able to compare speech utterances of a fixed length duration, which limits the ability of the model. Our model learns SV from speech of variable length.

Our work continues with the idea that a neural network-based TI SV model, pre-trained on large datasets, can be successfully applied for TI SV of speech-impaired patients in different languages. We evaluate the ability of TI SV systems to generalize by testing three models on three clinical trial datasets in five languages with variable length speech.

\section{Methods}

\subsection{Datasets}
In this paper, we used speech recordings from the three clinical trial studies (Table \ref{tab:speech-datasets}).

\begin{table}[tbp]
\centering
\setlength{\tabcolsep}{4pt} % Default value: 6pt
\footnotesize{  % Font size is decreased
  \begin{tabular*}{\columnwidth}{@{\extracolsep{\fill}} llllll}
      \hline
      \textbf{Language} & \textbf{English} & \textbf{German} & \textbf{Danish} & \textbf{Spanish} & \textbf{Arabic}\\
      \textbf{} & \textbf{(en)} & \textbf{(de)} & \textbf{(da)} & \textbf{(es)} & \textbf{(ar)}\\
      \hline
      \hline
      Dataset  & ADCT &  CSMCI & CSMCI & CSMCI & SCZCS\\
      \# Speakers & 659 &  29 & 69 & 43 & 30\\
      \# Samples & 7084 &  135 & 483 & 157 & 1192\\
      Avg. \# Samp. per Speaker & 10.7$\pm$7.0 & 4.7$\pm$2.5 & 7.0$\pm$3.6 & 3.6$\pm$1.0 & 39.7$\pm$12.0\\
      Avg. Dur. of Audio (sec) & 69.31 & 150.41 & 134.61 & 106.23 & 39.41 \\\
      Avg. Dur. of Speech (sec) & 37.30 & 110.07 & 89.31 & 74.46 & 21.88 \\
      \hline 
  \end{tabular*}
}
  \caption{Speech datasets used in this work. `\# Speakers' denotes the number of speakers in each dataset. `\# Samples' denotes the number of audio samples in each dataset.}
  \label{tab:speech-datasets}
\end{table}

\textbf{Alzheimer's Disease Clinical Trial (ADCT)} is a dataset that was collected during a clinical trial involving patients with mild to moderate Alzheimer's disease (AD). This is a longitudinal dataset of speech recordings of English speakers performing a picture description task \cite{goodglass2001bdae, becker1994natural}, as well as phonemic verbal fluency \cite{borkowski1967word} and semantic (categorical) verbal fluency \cite{tombaugh1999normative} tasks every 12 weeks for their 48-week treatment period. All participants were confirmed to have a clinical diagnosis of AD based on the National Institute on Aging/Alzheimer's Association (NIA-AA) criteria \cite{frisoni2011revised}.

\textbf{Clinical Study of Mild Cognitive Impairment (CSMCI)} is a dataset that was collected during the clinical trial involving patients with mild cognitive impairment (MCI) and early AD. This is a longitudinal dataset of speech recordings of German, Danish, and Spanish speakers performing picture description tasks \cite{goodglass2001bdae, becker1994natural} every 12 weeks for the 96-week treatment period. All participants have a clinical diagnosis of either MCI or mild AD according to NIA-AA criteria \cite{frisoni2011revised}.

\textbf{Schizophrenia Clinical Study (SCZCS)} is a dataset that was collected during the clinical trial involving patients with a diagnosis of schizophrenia (SCZ) based on diagnostic and statistical manual of mental disorders, fifth edition (DSM-5) \cite{american2013diagnostic}. This is a longitudinal dataset of speech recordings of Arabic speakers performing picture description  \cite{goodglass2001bdae, becker1994natural}, phonemic fluency, semantic fluency, paragraph reading \& recall, and journaling tasks. These tasks were conducted on a monthly basis for a duration of 6 months.

% All participants have a clinical diagnosis of either MCI or mild AD according to NIA-AA criteria \cite{frisoni2011revised}. 

There are substantially more samples and more speakers in the ADCT dataset than in CSMCI and SCZCS datasets (Table~\ref{tab:speech-datasets}). The SCZCS dataset has the most samples per speaker, with the smallest average duration of speech.  The CSMCI dataset has the longest average duration of speech, with the smallest average number of samples per speaker.

\subsection{Speech tasks}
In each trial, the subjects carried out a set of standardized speech tasks in every recording session. These speech tasks were used in multiple previous studies~\cite{zhu2018semi,eyre-etal-2020-fantastic,novikova-2021-robustness,tasnim-etal-2022-depac} due to the ability to generate speech patterns that could be examined for acoustic and linguistic characteristics associated with their mental and cognitive health condition:

\begin{itemize}
  \setlength{\itemsep}{1pt}
  \setlength{\parskip}{0pt}
  \setlength{\parsep}{0pt}
\item \textbf{Picture Description Task:}
The subject was presented with a static image depicting an event and was then asked to describe the scenario in their own words. Such tasks have been demonstrated to serve as reliable substitutes for spontaneous discourse \cite{giles1996performance}. Describing a picture was determined to be an effective speech task for eliciting situations requiring a higher level of cognitive effort and resulting in noticeable changes in speech, which can then be utilized to identify cognitive disorders such as AD or MCI \cite{mueller2018connected}. In all studies, proprietary images were employed. They were designed to match the style and content of the well-researched `Cookie theft' picture \cite{goodglass2001bdae}. The guiding principles utilized to develop these pictures were based on the core design guidelines described in \cite{patel2014park}.

\item \textbf{Phonemic Fluency Task:}
The FAS (`F', `A', `S') task \cite{borkowski1967word}, specifically focusing on the letter `F', was employed to evaluate phonemic verbal fluency in participants. Participants were asked to name as many unique words starting with the letter `F' as they can in one minute. This type of assessment has been extensively utilized in diverse populations, including individuals afflicted with AD \cite{crossley1997letter}.

\item \textbf{Semantic Fluency Task:}
To evaluate semantic (categorical) fluency, participants were asked to list as many different animals \cite{tombaugh1999normative}, household objects, or food items as they could think of in one minute. This assessment has been widely used in a variety of populations, including AD patients \cite{crossley1997letter}.

\item \textbf{Paragraph Reading \& Recall Task:} This task involves an individual reading a short story at the beginning of the session \cite{bransford1972contextual, newcomer1999glucose}. Participants were asked to read one of three standard paragraphs, each containing the same number of details and information content units. They were asked to recall information about the story just after reading and again at the end of the session. It is observed that people who have SCZ show deficiencies in memory recall \cite{Goldberg1989RecallMemory}, making these tasks good indicators for measuring SCZ.

\item \textbf{Journaling Task:}
Participants were provided with a prompt and asked to create narrative in response. Prompts are open-ended, allowing participants to provide as much or as little detail as they choose. An example prompt could be "what did you do yesterday". This task is used to assess the participants emotions, mental health, and verbal ability \cite{trifu2017linguistic, sohal2022efficacy}.

\end{itemize}

\subsection{Models}

In this study, we evaluated three state-of-the-art SV models (Table \ref{tab:models}), pre-trained on mix of languages, and evaluated on speech from speech-impaired patients speaking in English, German, Danish, Spanish, and Arabic languages in zero-shot settings. All models incorporate a combination of 1D convolution, batch normalization (BN), and Rectified Linear Unit (ReLU) to learn speech representation. We used NVIDIA NeMo (Neural Modules) toolkit \footnote{https://github.com/NVIDIA/NeMo} implementation of all models.

%\begin{figure}[h]
%    \centering
%    \includegraphics[width=\linewidth]{speakerNet.png}
%    \caption{SpeakerNet model architecture \cite{koluguri2020speakernet}} 
%    \label{fig:speakernet_diagram}
%\end{figure}

 The \textbf{SpeakerNet model} \cite{koluguri2020speakernet} %(Figure \ref{fig:speakernet_diagram}) 
 is based on the QuartzNet ASR (Automatic Speech Recognition) architecture, which consists of an encoder and a decoder. This model is smaller in size compared to the other two models. It consists of 5M parameters, making it very efficient at quickly generating speaker embeddings. It was trained on the VoxCeleb 1 \cite{nagrani2017voxceleb} and VoxCeleb 2 \cite{chung2018voxceleb2} datasets.

 The \textbf{TitaNet model} \cite{koluguri2022titanet} architecture is similar to the SpeakerNet's architecture but five times larger in terms of the number of parameters. It also uses squeeze excitation (SE) blocks and a global average pooling layer after each SE Block. Compared to the other two models, it shows the best reported EER on VoxCeleb1 \cite{nagrani2017voxceleb} dataset, but it is slower in generating speaker embeddings. This model was trained on the VoxCeleb 1, VoxCeleb 2 \cite{chung2018voxceleb2}, Fisher \cite{cieri2004fisher}, and Switch Board \cite{godfrey1993switchboard} datasets.

\begin{table}[t]
\centering
\setlength{\tabcolsep}{5pt}
\begin{tabular}{lll}
\hline
\textbf{Model} & \textbf{\# of Params} & \textbf{\# of Spkrs} \\
\hline
\hline
SpeakerNet & 5M & 7,205 \\
TitaNet & 25.3M & 16,681 \\
ECAPA-TDNN & 22M & 14,343 \\
\hline
\end{tabular}
\caption{Comparison of models, based on the model size. The number of parameters is represented in millions (M).}
\label{tab:models}
\end{table}

Finally, the \textbf{ECAPA-TDNN model} \cite{dawalatabad2021ecapa} is a time-delay neural network-based model that has been specifically designed for SV tasks. It uses SE blocks just like TitaNet does. The ECAPA-TDNN model has been shown to achieve similar performance to TitaNet although it has fewer parameters and it was trained on fewer data than Titanet, specifically the subset of VoxCeleb 1 \cite{nagrani2017voxceleb}, VoxCeleb 2 \cite{chung2018voxceleb2}, Fisher \cite{cieri2004fisher}, and Switch Board \cite{godfrey1993switchboard} datasets.

We separately evaluated each SV model on ADCT (en), CSMCI (es), CSMCI (de), and CSMCI (da), and SCZCS (ar) datasets. First, we generated embeddings of all audio files in each dataset within the same language. Then, we joined those embeddings to create a set of positive and negative pair tuples. The positive tuples comprised pairs of embeddings, where enrollment and test speech came from the same speaker. There was $\sum_{i=1}^m {n_i \choose 2}$ combinations of positive tuples, where $m$ is the number of speakers and $n_i$ is the number of speech recordings for the i-th speaker within the same dataset and language. The negative tuples are pairs of embeddings of speech belonging to different speakers within the same dataset and language. There was $\frac{1}{2}\sum_{i=1}^m {n_i * (N-n_i)}$ negative pairs, where $N$ is the total number of all audio files within the same dataset and language.

Once we had all positive and negative pair tuples, we calculated the cosine similarity between the vector embeddings in each tuple. Then, we calculated EER by balancing the true positive rate and the true negative rate. If the cosine similarity was above this balance point, the tuple was predicted to belong to the same speaker. Otherwise, the two embeddings were predicted to belong to different individuals.

% TABLE TO COMPARE MODELS
\begin{table*}[t]
    \centering
\begin{tabular}{llllllll}
\hline
\multirow{2}{*}{\textbf{Models}} & \multicolumn{1}{l}{\textbf{English}} & \multicolumn{1}{l} {\textbf{Non-English}} & \multicolumn{5}{c}{\textbf{Performance on Different Languages (EER\%)}}\\
 & \textbf{Test Data} & \textbf{Test Data} &     \textbf{English}                  &    \textbf{German}  &   \textbf{Danish}    &  \textbf{Spanish}  & \textbf{Arabic} \\
\hline
\hline
SpeakerNet \cite{koluguri2020speakernet}  & ADCT & SCZCS and CSMCI  & 4.99 & 0.52  & 1.16  & 4.29 & 14.97      \\
TitaNet \cite{koluguri2022titanet}     & ADCT & SCZCS and CSMCI  & 3.10 & \textbf{0.50} & \textbf{0.57} & \textbf{0.67} & 9.42 \\
ECAPA-TDNN \cite{dawalatabad2021ecapa}  & ADCT & SCZCS and CSMCI  & \textbf{2.69}  & 0.58  & 1.30  & 1.20 & \textbf{8.26} \\
\hline
% % TI-Monolingual \cite{chojnacka2021speakerstew}   & Keywords + & Speech Queries & \textbf{1.13} & \textbf{0.96} & - & \textbf{0.82} & -\\
% \hline 
% Marija: removed previous line as I couldn't find it in original paper. Also, there is no proper comparison that would make sense between TI-Monolingual and our datasets
\end{tabular}
\caption{
Benchmarking SV models across English, German, Danish, Spanish, and Arabic languages based on EER\% metric. All models are evaluated with cosine similarity. The best (lowest) EER achieved on each dataset is represented in bold. The ADCT, SCZCS, and CSMCI datasets of speech-impaired patients are proprietary datasets we collected for testing the TI SV models.
}

\label{tab:results-table}
\end{table*}

\begin{table}[tbp]
\centering
\setlength{\tabcolsep}{4pt} % Adjusts the space between columns
\footnotesize{  % Decreased font size
  \begin{tabular*}{\columnwidth}{@{\extracolsep{\fill}} llllll}
      \hline
      \textbf{ } & \textbf{English} & \textbf{German} & \textbf{Danish} & \textbf{Spanish} & \textbf{Arabic}\\
      \textbf{} & \textbf{(en)} & \textbf{(de)} & \textbf{(da)} & \textbf{(es)} & \textbf{(ar)}\\
      \hline
      \hline
      English & 0.0 & 31.3 & 24.6 & 59.3 & 85.5\\
      %\hline 
       German & - & 0.0 & 28.3 & 56.8 & 76.3\\
      %\hline 
       Danish & - & - & 0.0& 51.4 & 84.9\\
      %\hline 
       Spanish & - & - & - & 0.0 & 76.6\\
      %\hline 
       Arabic & - & - & - & - & 0.0\\
      \hline 
  \end{tabular*}
}
\caption{The eLinguistics metric evaluating the proximity among the studied languages. A lower score indicates a higher degree of similarity and the metric has commutative property, making matrix above symmetric. We are showing just values above diagonal to improve clarity.}
\label{tab:elinguistic_values}
\end{table}

It should also be noted that there were around $m$ times as many negative tuples as there were positive tuples due to there being more ways of creating negative tuple pairs than positive pairs.

\subsection{Language Selection}

Datasets available to us are mainly consisting of European (Germanic and Romance) languages which are also more common in datasets used for pre-training of examined models. In addition, we examine performance on Arabic language, which belongs to Semitic language family and has different linguistic features and structures, including more complex grammar, additional sounds, and right-to-left script. Similar to English, Arabic is spoken across many countries, with distinct dialects. We are focusing on Jordanian Arabic due to current data availability. Jordanian Arabic has its own phonetic variations and pronunciation patterns that differ from other Arabic languages. It also has unique idioms, expressions, phrases, cultural influences, intonation, and accent. In \cite{Malikeh2023Factors}, it is shown that verifying English speakers with accent is easier.

The proximity scores based on the eLinguistics metric \cite{eLinguistics2023} give additional information on how linguistically close studied languages are to each other (Table \ref{tab:elinguistic_values}). For example, German and Danish have closer linguistic ties to English than Spanish does, while Arabic remains the most distinct.

\section{Results}

\subsection{Benchmarking Speaker Verification Models across Different Languages}

TitaNet performed the best on the German, Danish, and Spanish languages, with ECAPA-TDNN following closely behind, and SpeakerNet having the worst performance (Table \ref{tab:results-table}). The outcomes on the English and Arabic datasets were the best when using the ECAPA-TDNN model, with 2.69\% and 8.42\% EER, respectively. This results may be due to the fact that TitaNet and ECAPA-TDNN are larger models, pre-trained on a higher number of speakers than SpeakerNet (Table \ref{tab:models}). Also, TitaNet is flexible with variable lengths of speech recordings.

% \cite{poddar2018speaker} - SV length of utterance influences the result
% \cite{arasteh2022effect} - effect of speech task and pathology on SV

All tested models achieve better EER on European languages in zero-shot setting than on English language. While this may look like a surprising outcome, it was also observed by \cite{chojnacka2021speakerstew}. We hypothesize that the primary cause of this is the diversity of the ADCT English Test Dataset, which is significantly larger in terms of the number of speakers and samples. Additionally, it is the only dataset among those tested that includes a variety of speech tasks. To investigate these assumptions further, we examine the EER for different tasks (Table \ref{tab:speech-tasks}). Lastly, considering the widespread use of English globally and the resultant diversity among English speakers, we anticipate that increase in EER may be partially attributed to the dataset containing a considerable number of individuals for whom English is not the first language, or who speak English with a range of accents.

All tested models exhibit notably poorer performance on Arabic compared to European languages, which aligns with our expectations. Several potential reasons account for this behavior. 

First, there is the lexical and acoustic similarities between the source and target languages. As shown in Table \ref{tab:elinguistic_values}, languages like German, Danish and English are closer to each other in terms of linguistic proximity, which could partially account for better performance in models trained on these languages. On the other hand, Arabic, with its higher eLinguistics value, poses a significant challenge due to its linguistic dissimilarity from other languages. This might explain the observed drop in performance when dealing with Arabic, underlining the importance of considering linguistic distance in cross-lingual SV \cite{chiswick2008linguistic}. Second, the models have been exposed to substantially more data from the tested European languages and others similar to them (e.g., French, Italian). In contrast, Arabic data is less prevalent in the pre-training dataset and is linguistically distinct from most languages in the pre-trained dataset. Additionally, there could be unique clinical patterns of speech impairment in Arabic that negatively affect model performance. Further research is needed to examine this hypothesis across speech tasks and cognitive diseases. Despite this, it's crucial to highlight that the ECAPA-TDNN model still demonstrates commendable performance, with an EER of 8.26\%, making it a viable option for zero-shot Speaker Verification (SV) in Arabic. Nevertheless, for optimal performance, we recommend fine-tuning ECAPA-TDNN with additional Arabic data.

% \ME{Another potential reason is more diversity in the English dataset in terms of the number of speech tasks, speakers, and data collection settings.}

%Marija: removing this as I removed TI-Monolingual result
% The performance of the TI-Monolingual model \cite{chojnacka2021speakerstew} on Spanish and German data is comparable to tested multilingual models (see Table \ref{tab:results-table}). TitaNet outperforms TI-Monolingual on both Spanish and German data which can be due to size of models, or quality of data they were pre-trained on. In addition, the datasets used for training and evaluating the TI-Monolingual model were constrained to include `Hey Google' and `OK Google' keywords, suggesting that their results don't generalize to unconstrained speech structures. Moreover,

The performance of the TitaNet and ECAPA-TDNN models is also compared across different languages (Table \ref{tab:balanced-datasets}), but in those tests we controlled for confounding factors. We matched the number of speakers and average number of samples per speaker across the experiments and sampled the datasets. TitaNet is outperforming ECAPA-TDNN in German, Spanish, and Danish, with a slight drop in performance when compared to the results from the data with the original number of speakers (Table \ref{tab:balanced-datasets} vs table \ref{tab:results-table}). In addition, the SV performance on German, Spanish, and Danish are still better than the English datasets. This may be due to the differences in speech patterns between patients with MCI and AD \cite{MartinezNicolas2021}. 

\begin{table*}[h]
\centering
\begin{tabular}{lllllll}
\hline
\multirow{2}{*}{\textbf{Language}} & \multirow{2}{*}{\textbf{Models}} & \multirow{2}{*}{\textbf{EER(\%)}} & \multirow{2}{*}{\textbf{TPR | TNR}} & \multirow{2}{*}{\textbf{\#Spkrs}} & \multirow{2}{*}{\textbf{\#Smpls}} & \textbf{Avg \#Smpls} \\
 &  &  &  &  &  & \textbf{per Spkr} \\ \hline
\multirow{2}{*}{English} & TitaNet & \textbf{5.30} & 0.946 | 0.948 & \multirow{2}{*}{29} & \multirow{2}{*}{212} & \multirow{2}{*}{7.31 $\pm$ 1.86} \\
 & ECAPA-TDNN & \textbf{5.30} & 0.946 | 0.948 &  &  &  \\ \hline
\multirow{2}{*}{German} & TitaNet & \textbf{0.50} & 1.000 | 0.990 & \multirow{2}{*}{29} & \multirow{2}{*}{135} & \multirow{2}{*}{4.66 $\pm$ 2.54} \\
 & ECAPA-TDNN & 0.58 & 0.994 | 0.994 &  &  &  \\ \hline
\multirow{2}{*}{Danish} & TitaNet & \textbf{1.22} & 0.990 | 0.980 & \multirow{2}{*}{29} & \multirow{2}{*}{188} & \multirow{2}{*}{6.48 $\pm$ 3.45} \\
 & ECAPA-TDNN & 1.60 & 0.981 | 0.986 &  &  &  \\ \hline
\multirow{2}{*}{Spanish} & TitaNet & \textbf{0.81} & 0.990 | 0.994 & \multirow{2}{*}{29} & \multirow{2}{*}{109} & \multirow{2}{*}{3.76 $\pm$ 1.07} \\
 & ECAPA-TDNN & 1.46 & 0.985 | 0.986 &  &  &  \\ \hline
\end{tabular}
  \caption{
  Benchmarking SV models across different languages with a balanced number of subjects per dataset. The best (lowest) EER\% for each language is bold. `\#Spkrs' denotes the number of speakers participating in each dataset. `\#Smpls' denotes the number of audio recordings per dataset.
  }
  \label{tab:balanced-datasets}
\end{table*}

% TABLE FOR COMPARING ON DIFFERENT TASKS
\begin{table*}[htbp]
\centering
\begin{tabular}{llllllll}
\hline
\multirow{2}{*}{\textbf{Speech Task}} & \multirow{2}{*}{\textbf{EER(\%)}} & \multirow{2}{*}{\textbf{TPR | TNR}} & \multirow{2}{*}{\textbf{\#Spkrs}} & \multirow{2}{*}{\textbf{\#Smpls}} & \textbf{Avg \#Smpls} & \textbf{Avg Dur of} & \textbf{Avg Dur of} \\
 &  &  &  &  & \textbf{per Spkr} & \textbf{Audio (sec)} & \textbf{Speech (sec)} \\ \hline
Phonetic Fluency & 4.45 & 0.956 | 0.955 & 394 & 1641 & 4.24 $\pm$ 0.94 & 49.32 $\pm$ 10.51 & 19.02 $\pm$ 9.26 \\
Semantic Fluency & 3.99 & 0.957 | 0.963 & 394 & 1643 & 4.25 $\pm$ 0.95 & 51.58 $\pm$ 9.30 & 21.33 $\pm$ 9.89 \\
Picture Description & \textbf{2.11} & 0.978 | 0.980 & 397 & 2932 & 7.54 $\pm$ 1.89 & 93.21 $\pm$ 54.61 & 58.82 $\pm$ 41.81 \\ \hline
\end{tabular}
  \caption{
  Benchmarking TitaNet SV model across different speech tasks on the ADCT English dataset. `\#Spkrs' denotes the number of speakers participating in each speech task. `\#Smpls' denotes the number of audio recordings per task.
  }
  \label{tab:speech-tasks}
\end{table*}

% Additionally, while the models perform worse on the European datasets when number of speakers and samples is reduced, the performance on Arabic language is better.

% NOTE: These claims are not that strong 
% Firstly, these results suggest that decreasing the number of speakers degrades the performance of the models, which is in line with literature \cite{arasteh2022effect}.

% Secondly, given that the number of speakers and type of cognitive tasks are the same, the difference in performance between English and non-English data can not be due to variations in cognitive tasks or the number of speakers in each trial. 

% However, the average length of audio and speech is higher in CSMCI compared to the ADCT dataset (Table \ref{tab:speech-datasets}), suggesting that the models are utilizing longer speech recordings to improve performance in determining the similarity between the speakers as confirmed by literature \cite{poddar2018speaker}. 

There can be many potential factors influencing the SV results across different datasets or languages. In particular, we will discuss the effect of three factors on SV performance including type of the mental disorder or cognitive impairment, type of the target language, and data collection procedure of a dataset.

Type of the mental disorder or cognitive impairment may impact the ability to distinguish participants from one another. Participants in the SCZCS are diagnosed with schizophrenia, whereas ADCT and CSMCI patients have Alzheimer's disease or MCI. It is known that the nature of speech may vary between AD, MCI, and schizophrenia \cite{MartinezNicolas2021}. 

Another potential factor influencing the ability to generalize across different languages, can be the lexical or acoustic similarities \cite{chiswick2008linguistic} between the source and target languages. Languages, which have similar vocabulary or linguistic characteristics to the source language may benefit when performing TI SV. If the source language and the target languages are dissimilar, this may influence the ability for the monolingual to generalize to that target language. We believe this is the reason behind worse performance on Arabic language \cite{chiswick2008linguistic}.

Finally, data collection procedure can impact the quality and length of audio and consequently SV performance. The three models we used in this study are able to handle variable lengths of speech but it is not known what length of speech is optimal. In addition, the influence of environmental noise and possible parts of examiner's speech is not taken into consideration when analysing the results.

\subsection{Evaluation} 

When observing the applicability of these models for our primary use case, we see that the usage of TI SV models is a feasible approach that could be applied in detecting duplicate participants across studies. However, setting a language-specific threshold may be necessary to get the best results. In our experiments, different tuned thresholds were used to maximize EER for each language. This may be caused by dialect variability within languages and amount of non-native speakers. It is also evident that these models seem to be sufficiently generalizable across languages without language-specific fine-tuning.

\subsection{Benchmarking Speaker Verification Models across Different Speech Tasks}

The TitaNet model is tested on ADCT (en) data across different speech tasks in Table \ref{tab:speech-tasks}. The picture description shows the lowest EER compared to the phonetic and semantic fluency tasks. These results suggests that picture description samples perform better in characterizing the differences between the speakers in SV tasks used in clinical settings. It is also important to note that average number of samples per speaker, and average duration of audio and speech are also the best for picture description task. In the future work, we will examine in more details influence of each of these confounding factors.

% \subsection{Deployment of Model}
% While we have demonstrated the potential for applying TI SV monolingual modals on clinical datasets across multiple languages, we have only applied this model in a development setting. Before using the model in a production environment, more work needs to be done to account for cofounding factors mentioned in previous subsection, as well as influence of accents and dialects within language on the performance. In addition, more feasibility tests should be done to look at the implications of deploying such a model in ethical and safe way.

\section{Conclusion}

This research paper proposes the utilization of speech recordings for speaker verification in clinical trials, particularly for detecting and monitoring cognitive and mental health disorders. We demonstrated effectiveness of TitaNet, ECAPA-TDNN, and SpeakerNet models in zero-shot settings on speech-impaired patients speaking European languages and Arabic. 

Our findings indicate that these TI SV models offer a promising solution to address the issue of duplicate participants without requiring fine-tuning, achieving high performance across multiple languages. We also show that the type of speech tasks conducted impacts the performance of SV models. This insight can inform the design and selection of appropriate tasks to enhance the accuracy and reliability of speaker verification systems.

In the future work, it would be valuable to establish a baseline by evaluating results from speakers without cognitive impairments in each language, providing a comparative measure of performance. Additionally, investigating the influence of specific diseases, noise, speech length, and the number of samples on SV model performance would deepen our understanding of the factors that affect accuracy.

Expanding the evaluation to include additional target languages would further validate the robustness and applicability of the SV models in diverse linguistic contexts. Lastly, exploring within-language performance variations, particularly in languages with distinct dialects, such as Arabic, would provide insights into the challenges and opportunities associated with dialectal variations in speaker verification.

\section{Ethical Considerations}
As with any research involving human subjects, ethical considerations must be taken into account when analyzing and reporting the results. In the context of the SV research, the following ethical considerations are relevant.

\subsection{Privacy}

Speaker verification systems rely on biometric data, which can raise privacy concerns. It is crucial to ensure that the privacy of the enrolled speakers and their sensitive personal information is protected throughout the process. Our research used clinical trial datasets containing recordings from patients with Alzheimer's disease and mild cognitive impairment and schizophrenia. Due to the sensitivity of the data, we have made sure to conceal the identities of the individuals in the study and refrain from using personally-identifiable information.

\subsection{Bias}
Speaker verification systems have the potential to reinforce biases in society if not properly designed and evaluated. For instance, if the training data is not representative of the diverse population, the resulting system may not perform equally well across all groups. We can't ensure the diversity of data that was used in the training of the model as we do not have access to the demographic characteristics of the datasets and other sensitive information of participants enrolled in the trials. %no work was done in this area.

\subsection{Misuse}

Speaker verification systems can be misused for unethical purposes, such as identity theft or surveillance. It is important to emphasize that our research is focused on developing and evaluating SV models for legitimate purposes, i.e., solving the duplicate participant problem.

\section{Limitations}

Applying speaker verification (SV) to cognitive and mental health data comes with certain domain-specific limitations. Firstly, the \textbf{variability in speech patterns} among individuals with cognitive and mental health disorders can lead to variations in speech characteristics and affect the performance of SV models, as they are typically trained on relatively standardized speech data. Additionally, the cognitive impairment manifestation of the same disease may vary with language and those patterns are understudied for non-English languages. The impairment patterns are also shown to be different for different diseases. Consequently, the EER of SV systems may be different when applied to populations with cognitive and mental health disorders, which is examined in \cite{Malikeh2023Factors}.

Another limitation arises from the \textbf{limited availability of data}. Obtaining large amounts of speech data from individuals with specific cognitive and mental health disorders is challenging. Datasets contain different amount of speakers for each severity level which impact is shown in \cite{Malikeh2023Factors}. Additionally, the \textbf{type of speech tasks} performed by individuals in clinical trials can influence speech characteristics and, subsequently, SV model performance as discussed in this paper. It is also important to consider the presence of \textbf{co-occurring disorders} among individuals with cognitive and mental health disorders. The presence of comorbid conditions can introduce additional variability in speech patterns, making it harder for SV models to accurately distinguish between speakers.

\bibliographystyle{ACM-Reference-Format}
\bibliography{literature}


\begin{thebibliography}{45}

%%% the Web address.

\ifx \showCODEN    \undefined \def \showCODEN     #1{\unskip}     \fi
\ifx \showDOI      \undefined \def \showDOI       #1{#1}\fi
\ifx \showISBNx    \undefined \def \showISBNx     #1{\unskip}     \fi
\ifx \showISBNxiii \undefined \def \showISBNxiii  #1{\unskip}     \fi
\ifx \showISSN     \undefined \def \showISSN      #1{\unskip}     \fi
\ifx \showLCCN     \undefined \def \showLCCN      #1{\unskip}     \fi
\ifx \shownote     \undefined \def \shownote      #1{#1}          \fi
\ifx \showarticletitle \undefined \def \showarticletitle #1{#1}   \fi
\ifx \showURL      \undefined \def \showURL       {\relax}        \fi
% invisible to TeX
\providecommand\bibfield[2]{#2}
\providecommand\bibinfo[2]{#2}
\providecommand\natexlab[1]{#1}
\providecommand\showeprint[2][]{arXiv:#2}

\bibitem[American Psychiatric~Association et~al\mbox{.}(2013)]%
        {american2013diagnostic}
\bibfield{author}{\bibinfo{person}{D American Psychiatric~Association}, \bibinfo{person}{American~Psychiatric Association}, {et~al\mbox{.}}} \bibinfo{year}{2013}\natexlab{}.
\newblock \bibinfo{booktitle}{\emph{Diagnostic and statistical manual of mental disorders: DSM-5}}. Vol.~\bibinfo{volume}{5}.
\newblock \bibinfo{publisher}{American psychiatric association Washington, DC}.
\newblock


\bibitem[Arasteh et~al\mbox{.}(2022)]%
        {arasteh2022speech}
\bibfield{author}{\bibinfo{person}{Soroosh~Tayebi Arasteh}, \bibinfo{person}{Tobias Weise}, \bibinfo{person}{Maria Schuster}, \bibinfo{person}{Elmar N{\"o}th}, \bibinfo{person}{Andreas Maier}, {and} \bibinfo{person}{Seung~Hee Yang}.} \bibinfo{year}{2022}\natexlab{}.
\newblock \showarticletitle{Is Speech Pathology a Biomarker in Automatic Speaker Verification?}
\newblock \bibinfo{journal}{\emph{arXiv preprint arXiv:2204.06450}} (\bibinfo{year}{2022}).
\newblock


\bibitem[Beaufils and Tomin(2023)]%
        {eLinguistics2023}
\bibfield{author}{\bibinfo{person}{Vincent Beaufils} {and} \bibinfo{person}{Johannes Tomin}.} \bibinfo{year}{2023}\natexlab{}.
\newblock \bibinfo{title}{Stochastic approach to worldwide language classification: the signals and the noise towards long-range exploration}.
\newblock
\newblock
\urldef\tempurl%
\url{http://www.eLinguistics.net}
\showURL{%
\tempurl}
\newblock
\shownote{Accessed: September 1, 2023}.


\bibitem[Becker et~al\mbox{.}(1994)]%
        {becker1994natural}
\bibfield{author}{\bibinfo{person}{James~T Becker}, \bibinfo{person}{Fran{\c{c}}ois Boiler}, \bibinfo{person}{Oscar~L Lopez}, \bibinfo{person}{Judith Saxton}, {and} \bibinfo{person}{Karen~L McGonigle}.} \bibinfo{year}{1994}\natexlab{}.
\newblock \showarticletitle{The natural history of Alzheimer's disease: description of study cohort and accuracy of diagnosis}.
\newblock \bibinfo{journal}{\emph{Archives of neurology}} \bibinfo{volume}{51}, \bibinfo{number}{6} (\bibinfo{year}{1994}), \bibinfo{pages}{585--594}.
\newblock


\bibitem[Borkowski et~al\mbox{.}(1967)]%
        {borkowski1967word}
\bibfield{author}{\bibinfo{person}{John~G Borkowski}, \bibinfo{person}{Arthur~L Benton}, {and} \bibinfo{person}{Otfried Spreen}.} \bibinfo{year}{1967}\natexlab{}.
\newblock \showarticletitle{Word fluency and brain damage}.
\newblock \bibinfo{journal}{\emph{Neuropsychologia}} \bibinfo{volume}{5}, \bibinfo{number}{2} (\bibinfo{year}{1967}), \bibinfo{pages}{135--140}.
\newblock


\bibitem[Bransford and Johnson(1972)]%
        {bransford1972contextual}
\bibfield{author}{\bibinfo{person}{John~D Bransford} {and} \bibinfo{person}{Marcia~K Johnson}.} \bibinfo{year}{1972}\natexlab{}.
\newblock \showarticletitle{Contextual prerequisites for understanding: Some investigations of comprehension and recall}.
\newblock \bibinfo{journal}{\emph{Journal of verbal learning and verbal behavior}} \bibinfo{volume}{11}, \bibinfo{number}{6} (\bibinfo{year}{1972}), \bibinfo{pages}{717--726}.
\newblock


\bibitem[Chen et~al\mbox{.}(2009)]%
        {chen2009speaker}
\bibfield{author}{\bibinfo{person}{Shi-Huang Chen}, \bibinfo{person}{Yu-Ren Luo}, {and} \bibinfo{person}{Rodrigo~Capobianco Guido}.} \bibinfo{year}{2009}\natexlab{}.
\newblock \showarticletitle{Speaker Verification Using Line Spectrum Frequency, Formant, and Support Vector Machine}. In \bibinfo{booktitle}{\emph{2009 11th IEEE International Symposium on Multimedia}}. IEEE, \bibinfo{pages}{562--566}.
\newblock


\bibitem[Cheng and Wang(2004)]%
        {cheng2004method}
\bibfield{author}{\bibinfo{person}{Jyh-Min Cheng} {and} \bibinfo{person}{Hsiao-Chuan Wang}.} \bibinfo{year}{2004}\natexlab{}.
\newblock \showarticletitle{A method of estimating the equal error rate for automatic speaker verification}. In \bibinfo{booktitle}{\emph{2004 International Symposium on Chinese Spoken Language Processing}}. IEEE, \bibinfo{pages}{285--288}.
\newblock


\bibitem[Chiswick and Miller(2008)]%
        {chiswick2008linguistic}
\bibfield{author}{\bibinfo{person}{Barry~R. Chiswick} {and} \bibinfo{person}{Paul~W. Miller}.} \bibinfo{year}{2008}\natexlab{}.
\newblock \showarticletitle{Linguistic Distance: A Quantitative Measure of the Distance Between English and Other Languages}.
\newblock \bibinfo{journal}{\emph{Journal of Multilingual and Multicultural Development}} (\bibinfo{year}{2008}), \bibinfo{pages}{1--11}.
\newblock


\bibitem[Chojnacka et~al\mbox{.}(2021)]%
        {chojnacka2021speakerstew}
\bibfield{author}{\bibinfo{person}{Roza Chojnacka}, \bibinfo{person}{Jason Pelecanos}, \bibinfo{person}{Quan Wang}, {and} \bibinfo{person}{Ignacio~Lopez Moreno}.} \bibinfo{year}{2021}\natexlab{}.
\newblock \showarticletitle{Speakerstew: Scaling to many languages with a triaged multilingual text-dependent and text-independent speaker verification system}.
\newblock \bibinfo{journal}{\emph{arXiv preprint arXiv:2104.02125}} (\bibinfo{year}{2021}).
\newblock


\bibitem[Chung et~al\mbox{.}(2018)]%
        {chung2018voxceleb2}
\bibfield{author}{\bibinfo{person}{Joon~Son Chung}, \bibinfo{person}{Arsha Nagrani}, {and} \bibinfo{person}{Andrew Zisserman}.} \bibinfo{year}{2018}\natexlab{}.
\newblock \showarticletitle{Voxceleb2: Deep speaker recognition}.
\newblock \bibinfo{journal}{\emph{arXiv preprint arXiv:1806.05622}} (\bibinfo{year}{2018}).
\newblock


\bibitem[Cieri et~al\mbox{.}(2004)]%
        {cieri2004fisher}
\bibfield{author}{\bibinfo{person}{Christopher Cieri}, \bibinfo{person}{David Miller}, {and} \bibinfo{person}{Kevin Walker}.} \bibinfo{year}{2004}\natexlab{}.
\newblock \showarticletitle{The Fisher corpus: A resource for the next generations of speech-to-text.}. In \bibinfo{booktitle}{\emph{LREC}}, Vol.~\bibinfo{volume}{4}. \bibinfo{pages}{69--71}.
\newblock


\bibitem[Crossley et~al\mbox{.}(1997)]%
        {crossley1997letter}
\bibfield{author}{\bibinfo{person}{Margaret Crossley}, \bibinfo{person}{Carl D'arcy}, {and} \bibinfo{person}{Nigel~SB Rawson}.} \bibinfo{year}{1997}\natexlab{}.
\newblock \showarticletitle{Letter and category fluency in community-dwelling Canadian seniors: A comparison of normal participants to those with dementia of the Alzheimer or vascular type}.
\newblock \bibinfo{journal}{\emph{Journal of clinical and experimental neuropsychology}} \bibinfo{volume}{19}, \bibinfo{number}{1} (\bibinfo{year}{1997}), \bibinfo{pages}{52--62}.
\newblock


\bibitem[Dawalatabad et~al\mbox{.}(2021)]%
        {dawalatabad2021ecapa}
\bibfield{author}{\bibinfo{person}{Nauman Dawalatabad}, \bibinfo{person}{Mirco Ravanelli}, \bibinfo{person}{Fran{\c{c}}ois Grondin}, \bibinfo{person}{Jenthe Thienpondt}, \bibinfo{person}{Brecht Desplanques}, {and} \bibinfo{person}{Hwidong Na}.} \bibinfo{year}{2021}\natexlab{}.
\newblock \showarticletitle{Ecapa-tdnn embeddings for speaker diarization}.
\newblock \bibinfo{journal}{\emph{arXiv preprint arXiv:2104.01466}} (\bibinfo{year}{2021}).
\newblock


\bibitem[Eyre et~al\mbox{.}(2020)]%
        {eyre-etal-2020-fantastic}
\bibfield{author}{\bibinfo{person}{Ben Eyre}, \bibinfo{person}{Aparna Balagopalan}, {and} \bibinfo{person}{Jekaterina Novikova}.} \bibinfo{year}{2020}\natexlab{}.
\newblock \showarticletitle{Fantastic Features and Where to Find Them: Detecting Cognitive Impairment with a Subsequence Classification Guided Approach}. In \bibinfo{booktitle}{\emph{Proceedings of the Sixth Workshop on Noisy User-generated Text (W-NUT 2020)}}. \bibinfo{publisher}{Association for Computational Linguistics}, \bibinfo{address}{Online}, \bibinfo{pages}{193--199}.
\newblock
\urldef\tempurl%
\url{https://doi.org/10.18653/v1/2020.wnut-1.25}
\showDOI{\tempurl}


\bibitem[Frisoni et~al\mbox{.}(2011)]%
        {frisoni2011revised}
\bibfield{author}{\bibinfo{person}{Giovanni~B Frisoni}, \bibinfo{person}{Bengt Winblad}, {and} \bibinfo{person}{John~T O'Brien}.} \bibinfo{year}{2011}\natexlab{}.
\newblock \showarticletitle{Revised {NIA-AA} criteria for the diagnosis of Alzheimer's disease: a step forward but not yet ready for widespread clinical use}.
\newblock \bibinfo{journal}{\emph{International psychogeriatrics}} \bibinfo{volume}{23}, \bibinfo{number}{8} (\bibinfo{year}{2011}), \bibinfo{pages}{1191--1196}.
\newblock


\bibitem[Giles et~al\mbox{.}(1996)]%
        {giles1996performance}
\bibfield{author}{\bibinfo{person}{Elaine Giles}, \bibinfo{person}{Karalyn Patterson}, {and} \bibinfo{person}{John~R Hodges}.} \bibinfo{year}{1996}\natexlab{}.
\newblock \showarticletitle{Performance on the Boston Cookie Theft picture description task in patients with early dementia of the Alzheimer's type: missing information}.
\newblock \bibinfo{journal}{\emph{Aphasiology}} \bibinfo{volume}{10}, \bibinfo{number}{4} (\bibinfo{year}{1996}), \bibinfo{pages}{395--408}.
\newblock


\bibitem[Godfrey and Holliman(1993)]%
        {godfrey1993switchboard}
\bibfield{author}{\bibinfo{person}{John Godfrey} {and} \bibinfo{person}{Edward Holliman}.} \bibinfo{year}{1993}\natexlab{}.
\newblock \showarticletitle{Switchboard-1 Release 2 LDC97S62}.
\newblock \bibinfo{journal}{\emph{Linguistic Data Consortium}} (\bibinfo{year}{1993}), \bibinfo{pages}{34}.
\newblock


\bibitem[Goldberg et~al\mbox{.}(1989)]%
        {Goldberg1989RecallMemory}
\bibfield{author}{\bibinfo{person}{TE Goldberg}, \bibinfo{person}{DR Weinberger}, \bibinfo{person}{NH Pliskin}, \bibinfo{person}{KF Berman}, {and} \bibinfo{person}{MH Podd}.} \bibinfo{year}{1989}\natexlab{}.
\newblock \showarticletitle{Recall memory deficit in schizophrenia: A possible manifestation of prefrontal dysfunction}.
\newblock \bibinfo{journal}{\emph{Schizophr Res}}  \bibinfo{volume}{2} (\bibinfo{year}{1989}), \bibinfo{pages}{251--257}.
\newblock
Issue 3.
\urldef\tempurl%
\url{https://doi.org/10.1016/0920-9964(89)90001-7}
\showDOI{\tempurl}


\bibitem[Goodglass et~al\mbox{.}(2001)]%
        {goodglass2001bdae}
\bibfield{author}{\bibinfo{person}{Harold Goodglass}, \bibinfo{person}{Edith Kaplan}, {and} \bibinfo{person}{Sandra Weintraub}.} \bibinfo{year}{2001}\natexlab{}.
\newblock \bibinfo{booktitle}{\emph{BDAE: The Boston diagnostic aphasia examination}}.
\newblock \bibinfo{publisher}{Lippincott Williams \& Wilkins Philadelphia, PA}.
\newblock


\bibitem[Irum and Salman(2019)]%
        {irum2019speaker}
\bibfield{author}{\bibinfo{person}{Amna Irum} {and} \bibinfo{person}{Ahmad Salman}.} \bibinfo{year}{2019}\natexlab{}.
\newblock \showarticletitle{Speaker verification using deep neural networks: A}.
\newblock \bibinfo{journal}{\emph{International Journal of Machine Learning and Computing}} \bibinfo{volume}{9}, \bibinfo{number}{1} (\bibinfo{year}{2019}).
\newblock


\bibitem[Kleynhans and Barnard(2005)]%
        {kleynhans2005language}
\bibfield{author}{\bibinfo{person}{Neil~T Kleynhans} {and} \bibinfo{person}{Etienne Barnard}.} \bibinfo{year}{2005}\natexlab{}.
\newblock \showarticletitle{Language dependence in multilingual speaker verification}.
\newblock  (\bibinfo{year}{2005}).
\newblock


\bibitem[Koluguri et~al\mbox{.}(2020)]%
        {koluguri2020speakernet}
\bibfield{author}{\bibinfo{person}{Nithin~Rao Koluguri}, \bibinfo{person}{Jason Li}, \bibinfo{person}{Vitaly Lavrukhin}, {and} \bibinfo{person}{Boris Ginsburg}.} \bibinfo{year}{2020}\natexlab{}.
\newblock \showarticletitle{SpeakerNet: 1D depth-wise separable convolutional network for text-independent speaker recognition and verification}.
\newblock \bibinfo{journal}{\emph{arXiv preprint arXiv:2010.12653}} (\bibinfo{year}{2020}).
\newblock


\bibitem[Koluguri et~al\mbox{.}(2022)]%
        {koluguri2022titanet}
\bibfield{author}{\bibinfo{person}{Nithin~Rao Koluguri}, \bibinfo{person}{Taejin Park}, {and} \bibinfo{person}{Boris Ginsburg}.} \bibinfo{year}{2022}\natexlab{}.
\newblock \showarticletitle{TitaNet: Neural Model for speaker representation with 1D Depth-wise separable convolutions and global context}. In \bibinfo{booktitle}{\emph{ICASSP 2022-2022 IEEE International Conference on Acoustics, Speech and Signal Processing (ICASSP)}}. IEEE, \bibinfo{pages}{8102--8106}.
\newblock


\bibitem[Li et~al\mbox{.}(2017)]%
        {li2017cross}
\bibfield{author}{\bibinfo{person}{Lantian Li}, \bibinfo{person}{Dong Wang}, \bibinfo{person}{Askar Rozi}, {and} \bibinfo{person}{Thomas~Fang Zheng}.} \bibinfo{year}{2017}\natexlab{}.
\newblock \showarticletitle{Cross-lingual speaker verification with deep feature learning}. In \bibinfo{booktitle}{\emph{2017 Asia-Pacific Signal and Information Processing Association Annual Summit and Conference (APSIPA ASC)}}. IEEE, \bibinfo{pages}{1040--1044}.
\newblock


\bibitem[Li et~al\mbox{.}(2022)]%
        {li2022interpretable}
\bibfield{author}{\bibinfo{person}{Xuhong Li}, \bibinfo{person}{Haoyi Xiong}, \bibinfo{person}{Xingjian Li}, \bibinfo{person}{Xuanyu Wu}, \bibinfo{person}{Xiao Zhang}, \bibinfo{person}{Ji Liu}, \bibinfo{person}{Jiang Bian}, {and} \bibinfo{person}{Dejing Dou}.} \bibinfo{year}{2022}\natexlab{}.
\newblock \showarticletitle{Interpretable deep learning: Interpretation, interpretability, trustworthiness, and beyond}.
\newblock \bibinfo{journal}{\emph{Knowledge and Information Systems}} \bibinfo{volume}{64}, \bibinfo{number}{12} (\bibinfo{year}{2022}), \bibinfo{pages}{3197--3234}.
\newblock


\bibitem[Malikeh et~al\mbox{.}(2023)]%
        {Malikeh2023Factors}
\bibfield{author}{\bibinfo{person}{Ehghaghi Malikeh}, \bibinfo{person}{Stanojevic Marija}, \bibinfo{person}{Akram Ali}, {and} \bibinfo{person}{Novikova Jekaterina}.} \bibinfo{year}{2023}\natexlab{}.
\newblock \showarticletitle{Factors Affecting the Performance of Automated Speaker Verification in Alzheimer’s Disease Clinical Trials}.
\newblock  (\bibinfo{year}{2023}).
\newblock


\bibitem[Martínez-Nicolás et~al\mbox{.}(2021)]%
        {MartinezNicolas2021}
\bibfield{author}{\bibinfo{person}{I. Martínez-Nicolás}, \bibinfo{person}{T.E. Llorente}, \bibinfo{person}{F. Martínez-Sánchez}, {and} \bibinfo{person}{J.J.G. Meilán}.} \bibinfo{year}{2021}\natexlab{}.
\newblock \showarticletitle{Ten Years of Research on Automatic Voice and Speech Analysis of People With Alzheimer's Disease and Mild Cognitive Impairment: A Systematic Review Article}.
\newblock \bibinfo{journal}{\emph{Frontiers in Psychology}}  \bibinfo{volume}{12} (\bibinfo{year}{2021}), \bibinfo{pages}{620251}.
\newblock
\urldef\tempurl%
\url{https://doi.org/10.3389/fpsyg.2021.620251}
\showDOI{\tempurl}


\bibitem[Melin(1996)]%
        {melin1996speaker}
\bibfield{author}{\bibinfo{person}{H{\aa}kan Melin}.} \bibinfo{year}{1996}\natexlab{}.
\newblock \showarticletitle{Speaker verification in telecommunication}.
\newblock \bibinfo{journal}{\emph{Department of Speech, Music and Hearing, KTH, Available from: http://www. speech. kth. se/\~{} melin/publications. html}} (\bibinfo{year}{1996}).
\newblock


\bibitem[Mueller et~al\mbox{.}(2018)]%
        {mueller2018connected}
\bibfield{author}{\bibinfo{person}{Kimberly~D Mueller}, \bibinfo{person}{Bruce Hermann}, \bibinfo{person}{Jonilda Mecollari}, {and} \bibinfo{person}{Lyn~S Turkstra}.} \bibinfo{year}{2018}\natexlab{}.
\newblock \showarticletitle{Connected speech and language in mild cognitive impairment and Alzheimer’s disease: A review of picture description tasks}.
\newblock \bibinfo{journal}{\emph{Journal of clinical and experimental neuropsychology}} \bibinfo{volume}{40}, \bibinfo{number}{9} (\bibinfo{year}{2018}), \bibinfo{pages}{917--939}.
\newblock


\bibitem[Nagrani et~al\mbox{.}(2017)]%
        {nagrani2017voxceleb}
\bibfield{author}{\bibinfo{person}{Arsha Nagrani}, \bibinfo{person}{Joon~Son Chung}, {and} \bibinfo{person}{Andrew Zisserman}.} \bibinfo{year}{2017}\natexlab{}.
\newblock \showarticletitle{Voxceleb: a large-scale speaker identification dataset}.
\newblock \bibinfo{journal}{\emph{arXiv preprint arXiv:1706.08612}} (\bibinfo{year}{2017}).
\newblock


\bibitem[Newcomer et~al\mbox{.}(1999)]%
        {newcomer1999glucose}
\bibfield{author}{\bibinfo{person}{John~W Newcomer}, \bibinfo{person}{Suzanne Craft}, \bibinfo{person}{Robert Fucetola}, \bibinfo{person}{Steven~O Moldin}, \bibinfo{person}{Gregg Selke}, \bibinfo{person}{Leilani Paras}, {and} \bibinfo{person}{Ryan Miller}.} \bibinfo{year}{1999}\natexlab{}.
\newblock \showarticletitle{Glucose-induced increase in memory performance in patients with schizophrenia}.
\newblock \bibinfo{journal}{\emph{Schizophrenia Bulletin}} \bibinfo{volume}{25}, \bibinfo{number}{2} (\bibinfo{year}{1999}), \bibinfo{pages}{321--335}.
\newblock


\bibitem[Novikova(2021)]%
        {novikova-2021-robustness}
\bibfield{author}{\bibinfo{person}{Jekaterina Novikova}.} \bibinfo{year}{2021}\natexlab{}.
\newblock \showarticletitle{Robustness and Sensitivity of {BERT} Models Predicting {A}lzheimer{'}s Disease from Text}. In \bibinfo{booktitle}{\emph{Proceedings of the Seventh Workshop on Noisy User-generated Text (W-NUT 2021)}}, \bibfield{editor}{\bibinfo{person}{Wei Xu}, \bibinfo{person}{Alan Ritter}, \bibinfo{person}{Tim Baldwin}, {and} \bibinfo{person}{Afshin Rahimi}} (Eds.). \bibinfo{publisher}{Association for Computational Linguistics}, \bibinfo{address}{Online}, \bibinfo{pages}{334--339}.
\newblock
\urldef\tempurl%
\url{https://doi.org/10.18653/v1/2021.wnut-1.37}
\showDOI{\tempurl}


\bibitem[{\"O}zbek et~al\mbox{.}(2018)]%
        {ozbek2018recognition}
\bibfield{author}{\bibinfo{person}{M~Erdal {\"O}zbek}, \bibinfo{person}{Mohamed~Amine Haytom}, {and} \bibinfo{person}{Estelle Cherrier}.} \bibinfo{year}{2018}\natexlab{}.
\newblock \showarticletitle{Recognition of biometric unlock pattern by GMM-UBM}. In \bibinfo{booktitle}{\emph{2018 26th Signal Processing and Communications Applications Conference (SIU)}}. IEEE, \bibinfo{pages}{1--4}.
\newblock


\bibitem[Patel and Connaghan(2014)]%
        {patel2014park}
\bibfield{author}{\bibinfo{person}{Rupal Patel} {and} \bibinfo{person}{Kathryn Connaghan}.} \bibinfo{year}{2014}\natexlab{}.
\newblock \showarticletitle{Park Play: A picture description task for assessing childhood motor speech disorders}.
\newblock \bibinfo{journal}{\emph{International Journal of Speech-Language Pathology}} \bibinfo{volume}{16}, \bibinfo{number}{4} (\bibinfo{year}{2014}), \bibinfo{pages}{337--343}.
\newblock


\bibitem[Ramya et~al\mbox{.}(2022)]%
        {ramya2022enhanced}
\bibfield{author}{\bibinfo{person}{R Ramya}, \bibinfo{person}{P~Deepan Nivash}, \bibinfo{person}{I~Anwar Khansa}, {and} \bibinfo{person}{B~Harish Iyappa}.} \bibinfo{year}{2022}\natexlab{}.
\newblock \showarticletitle{Enhanced Speaker Verification Incorporated with Face Recognition}. In \bibinfo{booktitle}{\emph{2022 3rd International Conference on Smart Electronics and Communication (ICOSEC)}}. IEEE, \bibinfo{pages}{1468--1471}.
\newblock


\bibitem[Shiovitz et~al\mbox{.}(2013)]%
        {shiovitz2013cns}
\bibfield{author}{\bibinfo{person}{Thomas~M Shiovitz}, \bibinfo{person}{Charles~S Wilcox}, \bibinfo{person}{Lilit Gevorgyan}, {and} \bibinfo{person}{Adnan Shawkat}.} \bibinfo{year}{2013}\natexlab{}.
\newblock \showarticletitle{CNS sites cooperate to detect duplicate subjects with a clinical trial subject registry}.
\newblock \bibinfo{journal}{\emph{Innovations in Clinical Neuroscience}} \bibinfo{volume}{10}, \bibinfo{number}{2} (\bibinfo{year}{2013}), \bibinfo{pages}{17}.
\newblock


\bibitem[Tamoto and Itou(2019)]%
        {tamoto2019voice}
\bibfield{author}{\bibinfo{person}{Atsuki Tamoto} {and} \bibinfo{person}{Katunobu Itou}.} \bibinfo{year}{2019}\natexlab{}.
\newblock \showarticletitle{Voice authentication by text dependent single utterance for in-car environment}. In \bibinfo{booktitle}{\emph{Proceedings of the 10th International Symposium on Information and Communication Technology}}. \bibinfo{pages}{336--341}.
\newblock


\bibitem[Tasnim et~al\mbox{.}(2022)]%
        {tasnim-etal-2022-depac}
\bibfield{author}{\bibinfo{person}{Mashrura Tasnim}, \bibinfo{person}{Malikeh Ehghaghi}, \bibinfo{person}{Brian Diep}, {and} \bibinfo{person}{Jekaterina Novikova}.} \bibinfo{year}{2022}\natexlab{}.
\newblock \showarticletitle{{DEPAC}: a Corpus for Depression and Anxiety Detection from Speech}. In \bibinfo{booktitle}{\emph{Proceedings of the Eighth Workshop on Computational Linguistics and Clinical Psychology}}, \bibfield{editor}{\bibinfo{person}{Ayah Zirikly}, \bibinfo{person}{Dana Atzil-Slonim}, \bibinfo{person}{Maria Liakata}, \bibinfo{person}{Steven Bedrick}, \bibinfo{person}{Bart Desmet}, \bibinfo{person}{Molly Ireland}, \bibinfo{person}{Andrew Lee}, \bibinfo{person}{Sean MacAvaney}, \bibinfo{person}{Matthew Purver}, \bibinfo{person}{Rebecca Resnik}, {and} \bibinfo{person}{Andrew Yates}} (Eds.). \bibinfo{publisher}{Association for Computational Linguistics}, \bibinfo{address}{Seattle, USA}, \bibinfo{pages}{1--16}.
\newblock
\urldef\tempurl%
\url{https://doi.org/10.18653/v1/2022.clpsych-1.1}
\showDOI{\tempurl}


\bibitem[Tombaugh et~al\mbox{.}(1999)]%
        {tombaugh1999normative}
\bibfield{author}{\bibinfo{person}{Tom~N Tombaugh}, \bibinfo{person}{Jean Kozak}, {and} \bibinfo{person}{Laura Rees}.} \bibinfo{year}{1999}\natexlab{}.
\newblock \showarticletitle{Normative data stratified by age and education for two measures of verbal fluency: FAS and animal naming}.
\newblock \bibinfo{journal}{\emph{Archives of clinical neuropsychology}} \bibinfo{volume}{14}, \bibinfo{number}{2} (\bibinfo{year}{1999}), \bibinfo{pages}{167--177}.
\newblock


\bibitem[Trifu et~al\mbox{.}(2017)]%
        {trifu2017linguistic}
\bibfield{author}{\bibinfo{person}{Raluca~Nicoleta Trifu}, \bibinfo{person}{Bogdan NEME{\c{S}}}, \bibinfo{person}{Carolina Bodea-Hațegan}, {and} \bibinfo{person}{Doina Cozman}.} \bibinfo{year}{2017}\natexlab{}.
\newblock \showarticletitle{Linguistic indicators of language in major depressive disorder (MDD). An evidence based research.}
\newblock \bibinfo{journal}{\emph{Journal of Evidence-Based Psychotherapies}} \bibinfo{volume}{17}, \bibinfo{number}{1} (\bibinfo{year}{2017}).
\newblock


\bibitem[Tu et~al\mbox{.}(2022)]%
        {tu2022survey}
\bibfield{author}{\bibinfo{person}{Youzhi Tu}, \bibinfo{person}{Weiwei Lin}, {and} \bibinfo{person}{Man-Wai Mak}.} \bibinfo{year}{2022}\natexlab{}.
\newblock \showarticletitle{A Survey on Text-Dependent and Text-Independent Speaker Verification}.
\newblock \bibinfo{journal}{\emph{IEEE Access}} (\bibinfo{year}{2022}).
\newblock


\bibitem[Upadhyay et~al\mbox{.}(2022)]%
        {upadhyay2022feature}
\bibfield{author}{\bibinfo{person}{Shrikant Upadhyay}, \bibinfo{person}{Mohit Kumar}, \bibinfo{person}{Ashwani Kumar}, \bibinfo{person}{Ramesh Karnati}, \bibinfo{person}{Gouse~Baig Mahommad}, \bibinfo{person}{Sara~A Althubiti}, \bibinfo{person}{Fayadh Alenezi}, {and} \bibinfo{person}{Kemal Polat}.} \bibinfo{year}{2022}\natexlab{}.
\newblock \showarticletitle{Feature Extraction Approach for Speaker Verification to Support Healthcare System Using Blockchain Security for Data Privacy}.
\newblock \bibinfo{journal}{\emph{Computational and Mathematical Methods in Medicine}}  \bibinfo{volume}{2022} (\bibinfo{year}{2022}).
\newblock


\bibitem[Zhou et~al\mbox{.}(2021)]%
        {zhou2021language}
\bibfield{author}{\bibinfo{person}{Yi Zhou}, \bibinfo{person}{Xiaohai Tian}, {and} \bibinfo{person}{Haizhou Li}.} \bibinfo{year}{2021}\natexlab{}.
\newblock \showarticletitle{Language agnostic speaker embedding for cross-lingual personalized speech generation}.
\newblock \bibinfo{journal}{\emph{IEEE/ACM Transactions on Audio, Speech, and Language Processing}}  \bibinfo{volume}{29} (\bibinfo{year}{2021}), \bibinfo{pages}{3427--3439}.
\newblock


\bibitem[Zhu et~al\mbox{.}(2018)]%
        {zhu2018semi}
\bibfield{author}{\bibinfo{person}{Zining Zhu}, \bibinfo{person}{Jekaterina Novikova}, {and} \bibinfo{person}{Frank Rudzicz}.} \bibinfo{year}{2018}\natexlab{}.
\newblock \showarticletitle{Semi-supervised classification by reaching consensus among modalities}. In \bibinfo{booktitle}{\emph{NeurIPS Workshop on Interpretability and Robustness in Audio, Speech, and Language IRASL}}.
\newblock


\end{thebibliography}

\appendix

\end{document}